\pdfoutput=1
\documentclass{article}

\PassOptionsToPackage{numbers,sort&compress}{natbib}
\usepackage[eandd, final]{neurips_2026}

\usepackage[utf8]{inputenc}
\usepackage[T1]{fontenc}
\usepackage{hyperref}
\usepackage{url}
\usepackage{booktabs}
\usepackage{amsmath,amssymb,amsfonts}
\usepackage{graphicx}
\usepackage{microtype}
\usepackage{xcolor}
\usepackage{array}
\usepackage{placeins}

\definecolor{myframepink}{RGB}{72,138,176}
\hypersetup{
    colorlinks=true,
    allcolors=myframepink,
    pdftitle={A Bounded Evaluation Protocol for Representational Systematicity},
    pdfauthor={Boyu Cao, Wenshuo Wang}
}

\newcommand{\ours}{\textsc{Bounded-RS}}
\newcommand{\StateUnsupported}{\emph{Unsupported}}
\newcommand{\StatePresentOnly}{\emph{Present-only}}
\newcommand{\StateCausalUse}{\emph{Causal-use}}
\newcommand{\StateUndetermined}{\emph{Undetermined}}
\newcommand{\Binding}{\textsc{Binding}}
\newcommand{\Geometry}{\textsc{Geometry}}
\newcommand{\score}{\operatorname{score}}
\newcommand{\sigmoid}{\operatorname{sigmoid}}

\title{A Bounded Evaluation Protocol for Representational Systematicity}

\author{%
Boyu Cao \qquad Wenshuo Wang\thanks{Corresponding author.} \\
School of Future Technology, South China University of Technology, China \\
\texttt{202364870251@mail.scut.edu.cn}
}

\begin{document}
\maketitle
\begingroup
\renewcommand{\thefootnote}{}
\footnotetext{\raggedright Code and data: \url{https://github.com/VinceWangVW/An-Evaluation-Protocol-for-Representational-Systematicity}.}
\endgroup

\begin{abstract}
Multiple studies have shown that neural networks can be behaviorally systematic, producing correct outputs for novel combinations of familiar parts. Whether they are also representationally systematic, that is, whether they represent these parts and their roles and use them to produce the outputs, remains hard to test, because each kind of evidence can mislead on its own: correct outputs can come from shortcuts, a decodable signal need not be used, an intervention can act through unintended pathways, and a failed search does not show absence. We propose a bounded evaluation protocol for representational systematicity (\ours{}): for an explanation that attributes a systematic behavior to a declared representation, it tests on a matched behavioral contrast whether that representation is recoverable and whether it is causally used, checks the validity of this evidence, and assigns the explanation one of four bounded states, \StateUnsupported{}, \StatePresentOnly{}, \StateCausalUse{}, or \StateUndetermined{}, through control-calibrated scores, gates, and stop rules. On a semantic-parsing testbed of 96 matched sets that we release, the tested role--filler \Binding{} explanation of fine-tuned T5 models is \StateCausalUse{} and the distributed \Geometry{} explanation is \StateUnsupported{} within its declared search space, verdicts that hold across calibration-compatible parameter configurations, training seeds, and T5-base. On 41 cases with known causal status, full adjudication licenses a false \StateCausalUse{} in 1 of 34 noncausal cases, against 13 when recoverability is taken as use, without missing any causal case. \ours{} thus turns the question of representational systematicity, debated since Fodor and Pylyshyn, into comparable and falsifiable verdicts on declared explanations, each reported with the evidence pattern that licenses it and, for a negative, the search space that bounds it.
\end{abstract}

\section{Introduction}
\label{sec:introduction}

Systematicity is the capacity to understand and produce novel combinations of familiar constituents by preserving their roles and relations: a semantic parser that maps ``John loves Mary'' to a meaning in which John is the lover should also map ``Mary loves John'' to the role-reversed meaning. A model can show this capacity in its behavior, by succeeding on held-out recombinations, without thereby representing the relevant constituents and roles and using them to produce its outputs; this second property, representational systematicity, is the stronger claim \citep{vegner2025behavioural}. Whether neural networks achieve systematicity through such representations has been debated since \citet{fodor1988connectionism} argued that connectionist models lack them, and the question has gained weight as neural networks reach human-like systematic behavior in controlled tasks \citep{lake2023human}. Answering it is hard because each kind of evidence fails in a characteristic way: correct outputs can come from shortcuts \citep{mccoy2019right,geirhos2020shortcut}, a signal decodable from hidden states need not be used by the model \citep{hewitt2019designing,ravichander2020probing,elazar2021amnesic}, an intervention can change outputs through off-manifold states or unintended pathways \citep{makelov2024subspace,zhang2024patching}, and a failed search shows only that the tested operations did not find a mechanism. A credible representation-level claim therefore requires several kinds of evidence and explicit rules that state which claim each pattern of evidence licenses.

Existing work on behavioral and representational systematicity supplies this evidence but usually leaves these rules implicit (Section~\ref{sec:related}). Behavioral benchmarks such as SCAN, CFQ, COGS, ReCOGS, and SLOG make the input--output contrast precise \citep{lake2018generalization,keysers2020measuring,kim2020cogs,wu2023recogs,li2023slog} but do not identify the representation behind success. Representation-level work proposes explanations of that success, such as role--filler binding \citep{smolensky1990tensor,mccoy2019rnns} and distributed geometric structure \citep{bernardi2020geometry,ito2022compositional}, and supplies tools to test them, such as probes, control tasks, interchange interventions, and activation patching \citep{alain2016understanding,hewitt2019designing,geiger2021causal,meng2022locating}. Each tool tests one link between a signal and behavior, and each explanation is usually tested on its own task and judged by its own standard, so it is left open which claim a given pattern of evidence licenses. The same evidence can then be reported as claims of very different strength, and verdicts on different explanations cannot be compared.

We close this gap with a bounded evaluation protocol for representational systematicity, \ours{}, which orchestrates these tools to adjudicate representation-level explanations of a fixed behavioral contrast (Section~\ref{sec:protocol}). Its data layer fixes the contrast with matched examples. Its evidence layer collects, for a declared explanatory family, a behavioral anchor, recoverability evidence, generation-time interchange, necessity, and repair interventions with matched controls, and realism and item-linkage checks. Its decision layer maps this evidence through control-calibrated scores, gates, and stop rules to one of four bounded states: \StateUnsupported{}, \StatePresentOnly{}, \StateCausalUse{}, or \StateUndetermined{}. A negative about a whole family is licensed only when every admissible object of a frozen closure basis fails; the basis is a declared set of object classes, and extending the negative beyond it requires validated detectors. Each design choice blocks one of the failure modes above, and the rules are frozen on calibration controls before any target outcome is interpreted, so \ours{} fixes in advance which claim, and at what scope, each pattern of evidence licenses.

We instantiate \ours{} on a controlled semantic-parsing testbed of 96 matched sets, whose seven items change one contrast at a time and provide exact interchange and repair pairs (Section~\ref{sec:evaluation_setup}). T5-small models \citep{raffel2020t5}, whose internals are fully accessible, are fine-tuned on 64 sets and evaluated on 32 held-out sets, which include all six sets of a construction absent from training. On this testbed we operationalize two explanations of systematic generalization as potentially coexisting families: \Binding{}, which locates the explanation in role--filler signals, and \Geometry{}, which locates it in distributed objects such as predicate-conditioned relation subspaces, cross-step bridges, and query--source relations. The two families share the behavioral anchor, state semantics, and claim rules, and each supplies its own detectors, interventions, and controls.

The results show that the two families receive different verdicts (Section~\ref{sec:results}). The tested \Binding{} instantiation is \StateCausalUse{}: its role--filler signal is recoverable beyond surface controls, and interventions on it change generation beyond matched controls. The declared \Geometry{} family is \StateUnsupported{} after closure: no admissible object in its frozen basis passes the causal-use (necessity) gate and no \StatePresentOnly{} reading holds, and since no geometry detector is validated as a truthful monitor, the verdict stays at the family level. Both verdicts hold across 487 calibration-compatible parameter configurations and five training seeds, matched-set resampling yields no positive \Geometry{} reversal, an engineered truth control shows that the cross-step \Geometry{} branch can return all three evidential states, both verdicts replicate on T5-base, and the \Binding{} chain returns \StateCausalUse{} in two of three seeds on a prespecified SLOG subdomain. Full adjudication is also more reliable than its component rules: on 41 cases with known causal status, it licenses \StateCausalUse{} falsely in 1 of 34 noncausal cases, against 13 when recoverability is taken as use and 7 when interventions are ungated, while retaining all 7 causal cases and abstaining on 6 of the 41.

In Section~\ref{sec:discussion} we argue that bounded adjudication matters beyond systematicity: representation-level claims become comparable and falsifiable once each is reported with the evidence pattern that licenses it and, for a negative, the search space that bounds it.

\section{Related Work}
\label{sec:related}

\paragraph{Behavioral systematicity.} Systematicity has mainly been evaluated through behavior: benchmarks withhold recombinations of familiar parts from training and score the outputs on them, and they differ in which parts they recombine. SCAN tests new compositions of familiar primitives in navigation commands \citep{lake2018generalization}; CFQ tests new compounds of familiar atoms in questions paired with database queries \citep{keysers2020measuring}; and COGS, its revision ReCOGS, and SLOG test familiar words in new syntactic roles and structures when sentences are parsed into logical forms, with SLOG focused on new structures \citep{kim2020cogs,wu2023recogs,li2023slog}. These benchmarks make the behavioral contrast precise, but a correct output is compatible with structured representations and with shortcuts alike \citep{vegner2025behavioural}, so they cannot show whether a model represents the constituents and their roles and uses them to produce its outputs.

\paragraph{Representational systematicity.} Work on representational systematicity proposes what produces systematic behavior and supplies tools to test such proposals. The proposals range from role--filler binding \citep{smolensky1990tensor,mccoy2019rnns,soulos2020discovering,feng2024binding} to distributed geometric structure \citep{bernardi2020geometry,ito2022compositional}. Probes test whether a variable is decodable from hidden states \citep{alain2016understanding,belinkov2022probing}, and control tasks and amnesic probing test whether the decoding reflects the representation and whether the model relies on it \citep{hewitt2019designing,elazar2021amnesic}. Intervention methods such as activation patching and interchange interventions test whether manipulating a hypothesized variable changes outputs \citep{vig2020investigating,meng2022locating,wang2023interpretability,geiger2021causal,geiger2024das}. Faithfulness work defines what a correct explanation must reflect \citep{jacovi2020faithfulness}, and interpretability benchmarks score how well these methods localize mechanisms, some against known circuits \citep{gupta2024interpbench,mueller2025mib}. In short, these tools each test one link between a signal and behavior, and each explanation is tested on its own task by its own standard, so they cannot say which representation-level claim their combined evidence licenses or compare explanations with each other. \ours{} instead adjudicates all these links for every explanation on one matched behavioral contrast under rules frozen in advance, and therefore yields what they cannot: comparable verdicts on different explanations, negatives bounded by a declared search space, and far fewer false causal claims than decodability or ungated interventions alone (Section~\ref{sec:detector_truthfulness_results}).

\section{The \texorpdfstring{\ours{}}{Bounded-RS} Protocol}
\label{sec:protocol}

\begin{figure}[t]
\centering
\includegraphics[width=0.95\linewidth]{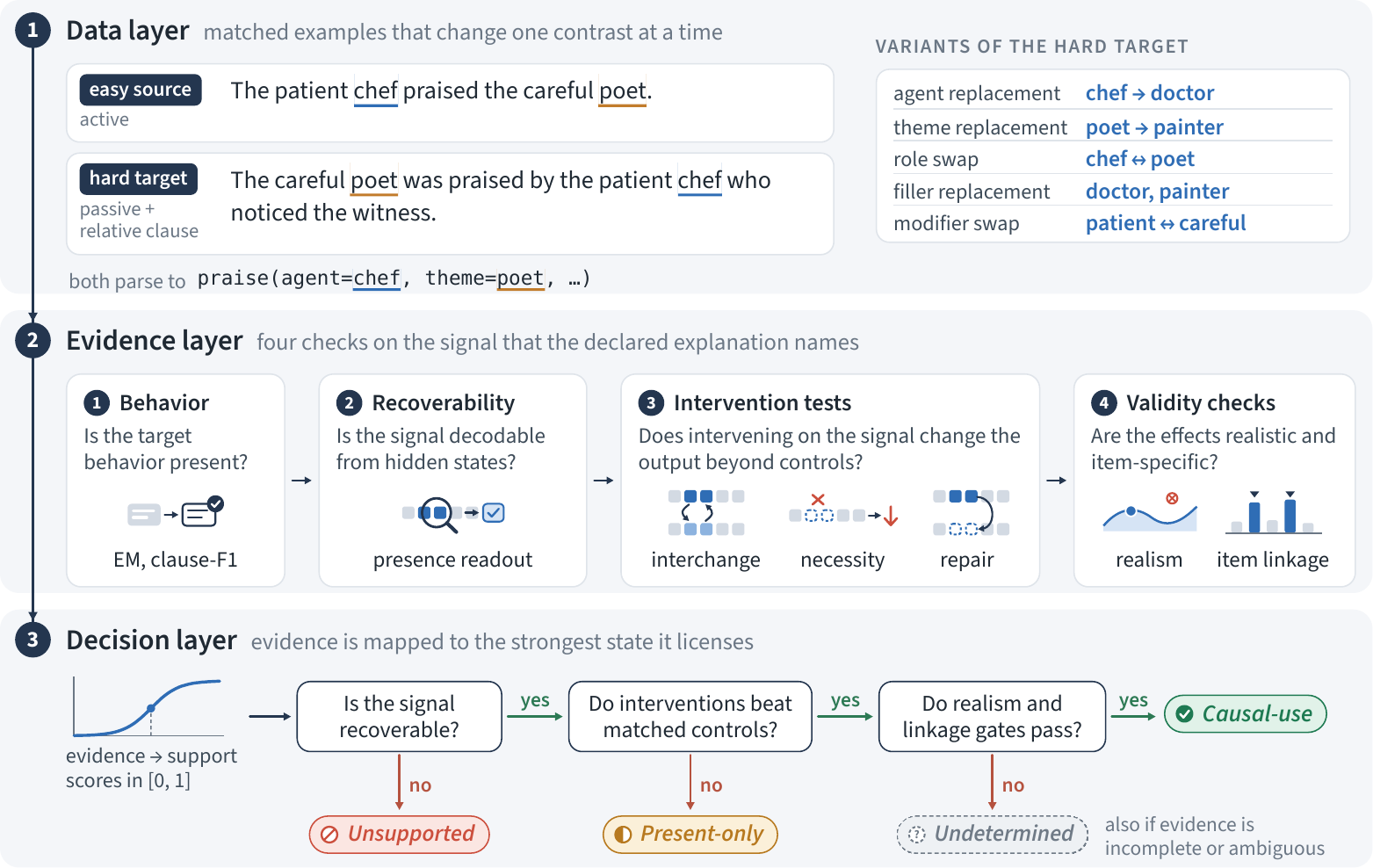}
\caption{Three-layer structure of \ours. The matched-set testbed defines controlled contrasts; family-specific evidence operations test recoverability and causal use; realism, linkage, and adjudication rules translate the evidence into bounded state decisions.}
\label{fig:three_layer_framework}
\end{figure}

\ours{} takes a behavioral systematicity result and a declared explanation of it and returns the strongest representation-level claim that the evidence licenses. Its inputs are a matched-set testbed for the behavioral contrast, a family specification that declares the hypothesized signal together with the operations that detect it, intervene on it, and could falsify it, and a control-calibrated decision rule; the testbed, state semantics, scoring grammar, gates, and stop rules are shared across families, whereas each family supplies its specification and the calibration controls that fit its boundaries (Appendix~\ref{app:scaffold_components}). Its output is a state together with the evidence rows that justify it. The protocol has three layers (Figure~\ref{fig:three_layer_framework}): the data layer fixes the behavioral contrast with matched examples (instantiated in Section~\ref{sec:controlled_testbed}), the evidence layer collects family-specific evidence about the hypothesized signal (Section~\ref{sec:evidence}), and the decision layer maps this evidence to a state (Section~\ref{sec:scoring}). Each design choice blocks one misreading from Section~\ref{sec:introduction}: behavior enters only as an anchor, so success alone cannot license use; recoverability and causal use are separate evidence roles, so a decodable signal is not called used; matched controls and realism and linkage gates keep effects of unintended pathways from counting as use; and closure over a frozen basis keeps a failed instantiation from being read as absence. We define the four states first, because they fix what the evidence and decision layers must establish.

\subsection{State semantics}
\label{sec:states}

Table~\ref{tab:states} defines the four states as evidential decisions under the declared operations. \StateUnsupported{} means that the tested operations did not support the signal or its use, not that no such representation exists. \StatePresentOnly{} names the case that probe-only studies leave open: the signal is recoverable, but causal-use evidence is insufficient. \StateUndetermined{} is an explicit abstention when the evidence does not license a stronger state (Section~\ref{sec:scoring}).

\begin{table}[t]
\centering
\small
\begin{tabular}{@{}p{0.18\linewidth}p{0.32\linewidth}p{0.38\linewidth}@{}}
\toprule
Decision & Minimum evidence pattern & Licensed claim \\
\midrule
\StateUnsupported{} & Weak, unstable, or unlinked presence evidence. & The protocol does not support the relevant representation or its causal use. \\
\StatePresentOnly{} & Positive recoverability evidence, with insufficient intervention, necessity, or repair evidence. & The representation is recoverable, but causal use is not supported. \\
\StateCausalUse{} & Recoverability plus controlled intervention, necessity, or repair evidence, with realism and linkage gates passing. & The representation is supported as present and causally used under the tested operations. \\
\StateUndetermined{} & Incomplete evidence, state-score margin below threshold, or failed or conflicting gates. & No stronger state decision is licensed. \\
\bottomrule
\end{tabular}
\caption{State decisions in \ours{}. Appendix~\ref{app:terminology} maps the state names to artifact labels.}
\label{tab:states}
\end{table}

\subsection{Evidence layer}
\label{sec:evidence}

Each evidence role rules out one alternative reading of the evidence (Table~\ref{tab:evidence_overview}). The behavioral anchor, exact match and clause-level F1 on the matched recombinations, establishes that the behavior to be explained is present. Recoverability tests whether the family's target signal is readable from hidden states beyond surface, filler, and predicate controls. Intervention tests act during generation: interchange transfers the signal between matched items, necessity removes or dampens it, and repair inserts or restores it from a matched easy source into a hard target with the same meaning (Section~\ref{sec:controlled_testbed}), and each effect is compared with the strongest of the family's matched controls, such as wrong-role, shuffled, or generic-decoder-gain controls. Validity checks reject effects that come from off-manifold intervention states (realism) or that appear only in aggregate rather than on the items the explanation predicts (item linkage). No single block is decisive; Appendix~\ref{app:evidence_blocks} defines each block and its metrics.

\begin{table}[t]
\centering
\small
\begin{tabular}{@{}p{0.20\linewidth}p{0.34\linewidth}p{0.34\linewidth}@{}}
\toprule
Evidence role & Blocks & What it rules out \\
\midrule
Behavioral anchor & Exact match and clause-level F1 on matched recombinations. & Representation-level claims when the target behavior is absent or unstable. \\
Recoverability & Family-specific presence readout with surface, filler, and predicate controls. & Claiming causal use for a signal that is not stably recoverable. \\
Intervention tests & Interchange, necessity, and repair interventions during generation. & Treating a merely decodable or correlational signal as used by the model. \\
Validity checks & Realism, item-linkage, shuffled, wrong-role, non-role, and generic-gain controls. & Accepting off-manifold interventions, aggregate-only effects, or generic decoder improvements. \\
\bottomrule
\end{tabular}
\caption{Evidence roles in \ours{} and the alternative reading each rules out.}
\label{tab:evidence_overview}
\end{table}

\subsection{Decision layer: scoring and adjudication}
\label{sec:scoring}

The decision layer first puts heterogeneous evidence on a common support scale. Each evidence block yields scalar features, and a raw feature $x$ with boundary $\tau$, scale $s$, and direction $d\in\{+1,-1\}$ is transformed as
\begin{equation}
\label{eq:score_transform}
\score(x;\tau,s,d)=\sigmoid\!\left(d\,\frac{x-\tau}{s}\right),
\end{equation}
so that a feature at its boundary scores 0.5 and $s$ sets how quickly support saturates. Group scores for behavior, presence, and use are weighted averages of their features, where required features must be present and optional ones are skipped when structurally unavailable; each state then receives a state score from how well the group scores match the evidence pattern it requires (Table~\ref{tab:states}), and realism, linkage, coverage, and margin act as gates. Behavior enters only as a prior with at most a 4\% share of the decision weight, so behavioral success alone cannot produce \StateCausalUse{}. All scoring parameters are fixed on calibration controls before target and closure outcomes are interpreted (Appendix~\ref{app:adjudicator_config}), and Appendix~\ref{app:walkthrough} follows one matched set through every step.

The decision layer then applies stop rules that decide which state the scores license. \StateCausalUse{} additionally requires the realism and linkage gates to pass, and \StateUndetermined{} is returned when evidence coverage does not exceed 0.80, when the top state leads the next by no more than 0.10, or when the gates that the top state requires fail or conflict. A positive state supports the tested family instantiation under its declared operations. A negative about a whole family needs more, because the space of instantiations is open: the family must freeze a closure basis, a declared set of object classes whose candidate objects are the admissible carriers of the signal, and every admissible object in it must fail the required criteria before a family-final \StateUnsupported{} is licensed. Extending such a negative to all mechanisms of the family's kind, an ontology-level claim, further requires detectors validated as truthful monitors under known positive, negative, and null conditions.

\section{Evaluation Setup: Testbed and Family Instantiations}
\label{sec:evaluation_setup}

\begin{figure}[t]
\centering
\includegraphics[width=0.95\linewidth]{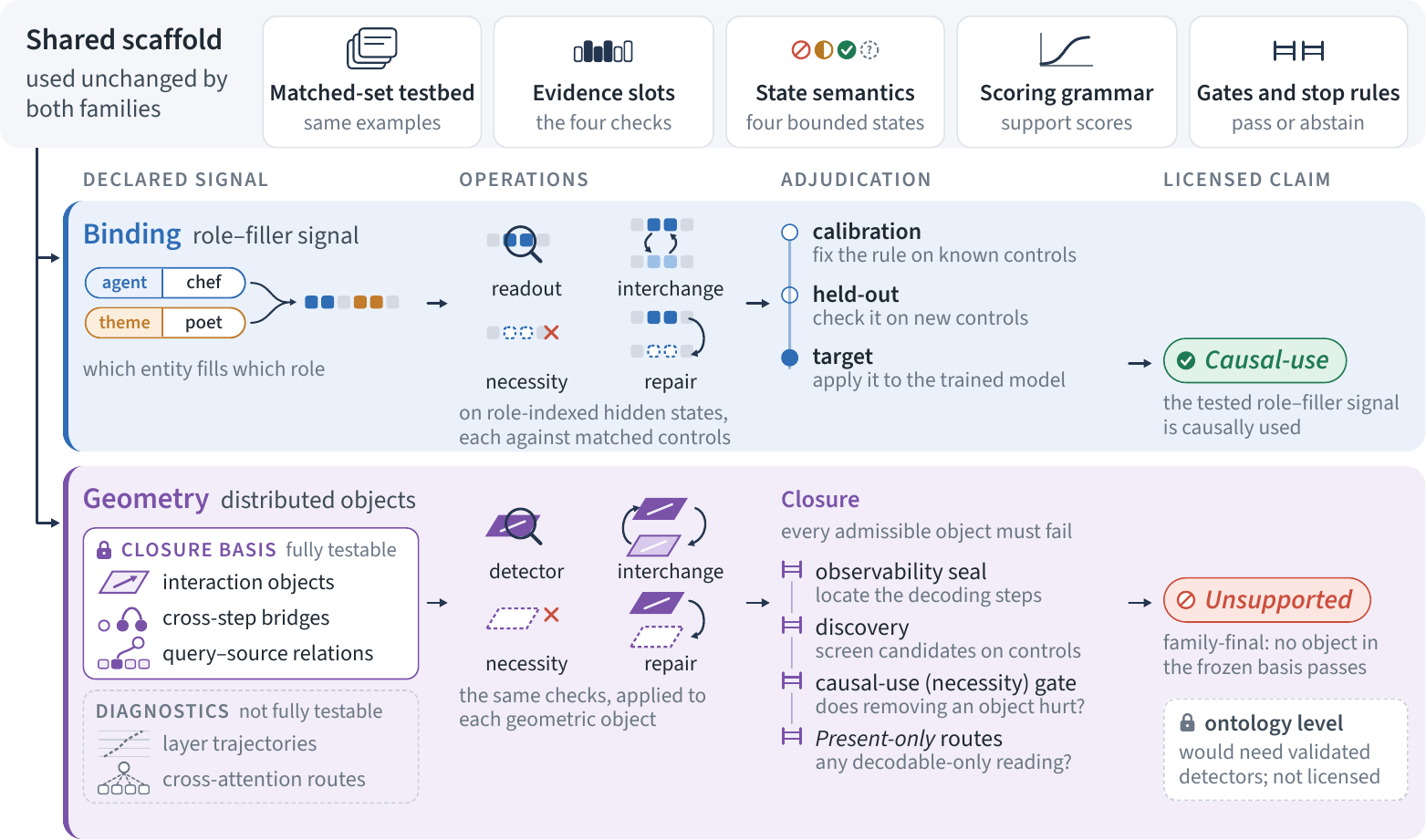}
\caption{Two-family audit workflow. Layer trajectories and routes are \Geometry{} diagnostics; the closure basis consists of interaction objects, cross-step bridges, and query--source relations. A family-final \Geometry{} negative requires closure, and an ontology-level negative would further require validated detectors.}
\label{fig:family_audit_flow}
\end{figure}

We instantiate the protocol for one behavioral contrast and two explanatory families, which may coexist in one model and therefore receive separate verdicts. This section describes the testbed and audited models (Section~\ref{sec:controlled_testbed}), the panels that calibrate and apply each decision rule (Section~\ref{sec:audit_panels}), and the two family instantiations (Sections~\ref{sec:binding_instantiation} and~\ref{sec:geometry_instantiation}); Figure~\ref{fig:family_audit_flow} summarizes the design.

\subsection{Controlled semantic-parsing testbed and audited models}
\label{sec:controlled_testbed}

The testbed is built for exact counterfactual comparison. Each matched set starts from one semantic design (predicate, agent, theme, alternate fillers, agent and theme modifiers, voice, and optional relative-clause context), whose meaning is written as a logical form such as \texttt{help(agent=freya,theme=kira,agent\_mod=brisk,theme\_mod=careful)}, and is rendered into seven items (Table~\ref{tab:matched_set}). A matched easy source in the active voice without embedding and a hard target in the set's construction share the gold meaning, which gives repair a source--target pair with aligned semantics; five further items each change one contrast, which supplies aligned pairs for interchange and controlled contrasts for the anchor, the readouts, and necessity. Each item is realized from validated templates (Appendix~\ref{app:data}), giving 672 items in 96 complete matched sets (28 active and 31 passive sets without embedding, 31 active and 6 passive sets with a relative clause), with 192 interchange pairs and 96 repair pairs, 68 of which have a hard target that differs from its source. All items are synthetic and contain no personal or sensitive information.

\begin{table}[t]
\centering
\small
\begin{tabular}{@{}p{0.21\linewidth}p{0.29\linewidth}p{0.37\linewidth}@{}}
\toprule
Matched item & Controlled relation & Protocol role \\
\midrule
Matched easy source & Same gold meaning as hard target; active voice without embedding. & Provides source signal for repair and matched behavior controls. \\
Hard target & Same gold meaning; the set's construction (passive voice, a relative clause, or neither). & Target for repair and failure analysis. \\
Agent replacement & Only the agent filler changes. & Tests role-specific sensitivity to who acts. \\
Theme replacement & Only the theme filler changes. & Tests role-specific sensitivity to who is acted on. \\
Role swap & Agent and theme exchange roles. & Tests role identity rather than name memorization. \\
Filler replacement & Both role fillers change. & Tests recombination under the same role structure. \\
Modifier swap & Agent and theme modifiers exchange attachment. & Tests modifier binding and attribution. \\
\bottomrule
\end{tabular}
\caption{The seven items of each matched set, shared by both families.}
\label{tab:matched_set}
\end{table}

We audit T5 models \citep{raffel2020t5} because their internals are fully accessible, which makes generation-time interventions tractable, and because their encoder--decoder split lets a control keep the encoder-side signal while weakening the decoder's access to it (Section~\ref{sec:audit_panels}). T5-small is fine-tuned on 64 matched sets (448 items) for 8 epochs with AdamW (learning rate $5\times10^{-5}$, batch size 8) and decodes greedily; the 32 held-out sets (224 items) form a development split of 128 items and a generalization split of 96 items (Appendix~\ref{app:data}). Passive sentences with a relative clause never occur in training, and all six such sets (42 items) are held out, so success requires recombining a voice and an embedding that were seen only separately. The normally fine-tuned target of the main panels reaches exact match and clause-level F1 of 1.0 on both held-out splits, which establishes the behavior whose representational basis is audited.

\subsection{Audit panels}
\label{sec:audit_panels}

Each family applies its frozen rule to calibration, held-out, and target panels, whose entries are runs: one T5 checkpoint, possibly modified, processed by that family's evidence operations to yield one state. Calibration controls are built to instantiate intended states: \StateCausalUse{} controls are trained on the full training split, \StatePresentOnly{} controls keep the encoder-side signal of such a checkpoint but weaken the decoder's access to it, for example by scaling decoder cross-attention to 0.25, and \StateUnsupported{} controls are trained on permuted targets or on a small subset of the training split. Calibration controls fix the boundaries and gates, and held-out controls built the same way from independent seeds test the frozen rule. Target panels contain the checkpoints to be adjudicated: the normally fine-tuned target that carries the claim and regimes built to probe the other states. Closure instead tests objects and serves only a family-level negative: once a closure basis is frozen, every admissible object in it is tested, so target states describe instantiations while closure decides the family. Beyond these panels, the two verdicts are tested with a parameter grid, additional training seeds, a matched-set bootstrap, engineered truth controls, and ports to T5-base and SLOG, and full adjudication is compared with its component rules on a separate panel (Appendix~\ref{app:result_details}, Table~\ref{tab:app_checks}).

\subsection{Binding instantiation}
\label{sec:binding_instantiation}

\Binding{} explains systematicity through role--filler assignments, that is, by keeping track of which entity fills the agent role and which fills the theme role across recombinations \citep{smolensky1990tensor,mccoy2019rnns}. Its target signal is a role-indexed hidden-state signal or an aggregate role--filler signal. Presence readouts recover role and filler information from hidden states; interchange and repair patch role-indexed signals across matched items, notably from the matched easy source into the hard target; and necessity removes or dampens role-relevant information and tests for selective degradation of the parse. Wrong-role, non-role, shuffled, path-only, and generic-decoder controls tie every effect to the declared role--filler operation, so an improvement on the hard target counts as evidence for binding only if it exceeds what these controls produce.

\subsection{Geometry instantiation and closure basis}
\label{sec:geometry_instantiation}

\Geometry{} explains systematicity through distributed structure that need not be localized in one role-indexed vector \citep{bernardi2020geometry,ito2022compositional}. Its operations target relation subspaces, layer trajectories, cross-attention routes, cross-step bridges, and query--source relations, and fill the same evidence slots as \Binding{} while manipulating geometric objects instead of a single vector. \Geometry{} is calibrated on its own control suite of independently trained checkpoints, so its calibration shares no checkpoint with \Binding{} (Appendix~\ref{app:families}); these controls calibrate the operation-level rule that precedes closure.

Because geometry-like explanations form an open-ended space, a negative about \Geometry{} needs a frozen closure basis (Section~\ref{sec:scoring}). An object class enters the basis only if it admits an observable object, a matched detector, object-specific intervention and repair, and falsifying controls. Three classes qualify: interaction objects, layer-local relation subspaces read conditionally on the predicate by a predicate-conditioned relation detector; cross-step bridges, read by a cross-step bridge detector based on cross-step alignment, with a max-statistic correction for the candidate search; and query--source relations, read by an argument-locus detector that relates a decoding query to the relevant source argument. Layer trajectories and routes lack a closure-grade repair and control chain and remain diagnostics, and nonlinear manifolds, higher-order or polysemantic interactions, sparse-feature circuits, diffuse long-range dynamic routes, and nonlocal multi-source mixtures lie outside the basis (Appendix~\ref{app:closure_basis}). Each class has 12 candidate objects built from templates that vary the head choice, the subspace basis, and the storage variant. Before closure, observability must be sealed, that is, a step aligner must locate on the controls the decoding steps at which objects are read. Discovery then keeps a candidate if its contrast on the calibration controls exceeds a minimum gap and accepts a class only if at least 8 of its 12 candidates survive, covering at least two decoding steps or step pairs and two templates; closure asks whether any object of an accepted class passes the causal-use (necessity) gate, and whether a \StatePresentOnly{} reading holds on the decoder-scaled, trained-scaled, and readout-masked \StatePresentOnly{} regimes of the control and target panels, over all items (the direct route) or within predeclared item slices (the slice-local route).

\section{Results}
\label{sec:results}

Section~\ref{sec:binding_results} adjudicates \Binding{}, Section~\ref{sec:geometry_results} adjudicates \Geometry{} and closes its basis, and Section~\ref{sec:detector_truthfulness_results} bounds the \Geometry{} negative, tests transfer to T5-base and SLOG, and compares full adjudication with its component rules. Table~\ref{tab:panels} summarizes the panels, and Appendix~\ref{app:result_details} reports them in detail.

\begin{table}[t]
\centering
\small
\begin{tabular}{@{}p{0.20\linewidth}p{0.11\linewidth}p{0.25\linewidth}p{0.32\linewidth}@{}}
\toprule
Panel & $n$ & Outcome & Diagnostic role \\
\midrule
\Binding{} calibration & 3 & all intended states & Controls separate the three evidential regimes. \\
\Binding{} held-out controls & 6 & acc.\ 0.833; macro-F1 0.822 & The frozen rule holds outside the calibration panel; one \StatePresentOnly{} control is escalated. \\
\Binding{} target audit & 5 & 1 \StateCausalUse{}, 1 \StatePresentOnly{}, 1 \StateUnsupported{}, 2 \StateUndetermined{} & The normally fine-tuned target is \StateCausalUse{}; other regimes are separated or withheld. \\
\Geometry{} held-out controls & 9 & acc.\ 0.778; macro-F1 0.796 & \StatePresentOnly{} is the least stable \Geometry{} state. \\
\Geometry{} target audit & 7 & 2 \StateCausalUse{}, 1 \StatePresentOnly{}, 3 \StateUnsupported{}, 1 \StateUndetermined{} & Operation-level states before closure; no family-level claim. \\
\Geometry{} closure & 36 objects in 3 classes & 0 objects pass closure & Licenses a family-final \StateUnsupported{} under the frozen basis. \\
Detector truthfulness & 3 detector classes & 0 validated detectors & Keeps the \Geometry{} verdict family-final. \\
\bottomrule
\end{tabular}
\caption{Panel-level results. Calibration outcomes, \Geometry{} variant counts, and detector-truthfulness checks are detailed in Appendix~\ref{app:result_details}; per-run measurements are in the released reports.}
\label{tab:panels}
\end{table}

\subsection{Binding audit}
\label{sec:binding_results}

We first check that the frozen \Binding{} rule recovers known states (Table~\ref{tab:panels}, first three rows). The three calibration controls receive their intended states, and on six held-out controls, two per state, the rule is correct on five (accuracy 0.833, macro-F1 0.822); the one error escalates a \StatePresentOnly{} control to \StateCausalUse{}. Removing the behavior features leaves both held-out metrics unchanged (Appendix~\ref{app:binding_result_details}). The rule thus holds outside its calibration panel, behavior decides no state, and its weak point is the \StatePresentOnly{} boundary.

We then apply the rule to the five-run target panel, which contains the normally fine-tuned target and four regimes built to probe the other states. The normally fine-tuned target is \StateCausalUse{}: its role signal is recoverable beyond surface controls, and its interchange, necessity, and repair effects exceed the strongest matched controls with realism and linkage gates passing. The target derived from it by scaling decoder cross-attention to 0.25 keeps a recoverable signal and becomes \StatePresentOnly{}, the target trained on 128 items is \StateUnsupported{}, and an instruction-prompted checkpoint with the same scaling and a target trained on longer surface realizations are \StateUndetermined{}, because their evidence does not separate any state clearly enough. The rule thus returns all four states, and only the normally fine-tuned target is \StateCausalUse{}.

Finally, we test whether this verdict depends on the frozen parameters, the training seed, or the sampled matched sets. It holds in all 487 calibration-compatible configurations, namely those of a 729-configuration grid over every verdict-relevant boundary, scale, weight, and gate that keep every calibration control in its intended state, and in five of five training seeds. For the first seed, the interchange advantage over the strongest control is 0.337 (95\% matched-set bootstrap interval [0.301, 0.371]) against the 0.245 use boundary, and the intervals for necessity, repair, realism, and linkage also lie on the supporting side of their decision boundaries (Appendix~\ref{app:robustness}). The tested \Binding{} instantiation is therefore \StateCausalUse{}: in these models, systematic behavior on the held-out matched sets is supported by a role--filler signal that is causally used during generation.

\subsection{Geometry audit}
\label{sec:geometry_results}

As in Section~\ref{sec:binding_results}, we first check the frozen \Geometry{} rule on held-out controls (Table~\ref{tab:panels}, rows 4--5). It recovers all six \StateUnsupported{} and \StateCausalUse{} controls but only one of three \StatePresentOnly{} controls (accuracy 0.778, macro-F1 0.796), so \StatePresentOnly{} is again its least reliable state.

Applied to the seven-run target panel, the rule gives operation-level states: the two fully trained targets are \StateCausalUse{}, three targets are \StateUnsupported{}, one is \StatePresentOnly{}, and one is \StateUndetermined{}, and five further variants built to produce \Geometry{} \StatePresentOnly{} regimes either collapse to \StateUnsupported{} or \StateCausalUse{} or retain a single \StatePresentOnly{} case (Appendix~\ref{app:geometry_variant_details}). These states show that a geometry transform changes generation but not which geometric object carries the change; the variant that attenuates a relation-bearing cross-attention subspace (Variant~D), for example, is \StateCausalUse{} on interchange evidence alone, with necessity tied to the strongest control and no repair evidence. Operation-level states therefore cannot support a family-level claim, and whether any geometric object carries a causal effect is decided by closure.

\begin{table}[t]
\centering
\small
\begin{tabular}{@{}p{0.23\linewidth}p{0.17\linewidth}p{0.09\linewidth}p{0.41\linewidth}@{}}
\toprule
Closure component & Candidates & Passed & Failure mode \\
\midrule
Observability seal & 1{,}536 trace rows & sealed & A1 selected; predicate-step coverage 1.0; argument steps not treated as validated object evidence. \\
Interaction objects & 12 objects; 4 survivors & 0 & Class not accepted at discovery, so no object can pass the causal-use (necessity) gate. \\
Cross-step bridges & 12 objects; 8 survivors & 0 & Class accepted; all 12 objects fail the causal-use (necessity) gate: gaps below threshold, null bands violated, realism below threshold. \\
Query--source relations & 12 objects; 4 survivors & 0 & Class not accepted at discovery, so no object can pass the causal-use (necessity) gate. \\
Direct \StatePresentOnly{} route & 8 object--regime rows & 0 & No global \StatePresentOnly{} route. \\
Slice-local \StatePresentOnly{} route & tested slices & 0 & No local slice passes. \\
\bottomrule
\end{tabular}
\caption{\Geometry{} closure audit. The verdict is family-final under the declared closure basis.}
\label{tab:closure}
\end{table}

Closure proceeds in the order of Table~\ref{tab:closure}. Observability is sealed first: the selected aligner, A1, covers the predicate step in all 1{,}536 control trace rows, while argument steps remain unaligned and are not treated as object evidence. Discovery then keeps 16 of the 36 candidate objects but accepts only the cross-step class, so interaction and query--source objects cannot pass the causal-use (necessity) gate, and all 12 cross-step objects fail it; neither the direct nor the slice-local \StatePresentOnly{} route holds. No admissible object of the frozen basis thus passes closure. This negative holds in all 487 calibration-compatible configurations and in five of five training seeds, and across 1{,}000 matched-set bootstrap replicates no object passes closure and the family is \StateUnsupported{} in 998 and \StateUndetermined{} in 2, never \StateCausalUse{} (Appendix~\ref{app:robustness}).

Finally, an engineered truth control checks that the cross-step branch, the only accepted class, can find a causal object when one exists. In in-basis triplets where a designated signal is causally used, equally recoverable but unused, or replaced by matched null content, blind discovery finds the designated object in all three causal seeds, and the branch returns \StateCausalUse{}, \StatePresentOnly{}, and \StateUnsupported{} correctly in 3/3 seeds each, with a correct-object advantage of 0.335 above the 0.245 use boundary (Appendix~\ref{app:truth_controls}). The negative is therefore not forced by the search, and the declared \Geometry{} family is \StateUnsupported{} at the family level: in these models, no admissible object of its frozen basis passes closure.

\subsection{Scope, transfer, and reliability of the verdicts}
\label{sec:detector_truthfulness_results}

In terms of scope, the \Geometry{} verdict stays at the family level. An ontology-level negative would require detectors validated as truthful monitors in trained models (Section~\ref{sec:scoring}), but none of the three detector classes passes this validation, each tested on 336 canonical examples (Appendix~\ref{app:detector_truthfulness_details}).

In terms of transfer, both verdicts carry over to T5-base, and the \Binding{} chain runs unchanged on a prespecified subdomain of an established benchmark. With the protocol and family interfaces unchanged, three T5-base seeds pass the behavioral anchor, held-out controls are correct in 16 of 18 \Binding{} and 22 of 27 \Geometry{} cases, and under both the frozen T5-small rules and rules recalibrated on T5-base controls, all six seed-by-rule runs return \Binding{} \StateCausalUse{} and \Geometry{} family-final \StateUnsupported{}. On a 96-pair exact-contrast subdomain of SLOG \citep{li2023slog}, selected before outcomes were viewed, the unchanged \Binding{} chain returns \StateCausalUse{}, \StateUndetermined{}, and \StateCausalUse{} for three T5-small seeds (Appendix~\ref{app:transfer}).

In terms of reliability, full adjudication makes far fewer false causal claims than its component rules. On a frozen panel of 41 predeclared cases with known causal status, taking recoverability as evidence of use licenses \StateCausalUse{} falsely in 13 of 34 noncausal cases, and licensing it from ungated interventions does so in 7; full \ours{} does so in 1, retains all 7 causal cases, and abstains on 6 of the 41 (Appendix~\ref{app:component_rules}).

\section{Discussion}
\label{sec:discussion}

We introduced \ours{}, which adjudicates representation-level explanations of a behavioral systematicity result into four bounded states, and applied it to two explanatory families on a controlled semantic-parsing testbed. Applied to fine-tuned T5-small and T5-base models, it licenses \StateCausalUse{} for the tested role--filler \Binding{} instantiation and a family-final \StateUnsupported{} for the declared \Geometry{} family, and both verdicts hold across calibration-compatible parameter configurations and training seeds, with no positive \Geometry{} reversal under matched-set resampling. The same discipline applies wherever internal representations are invoked to explain behavior. A probing study that decodes role information would report \StatePresentOnly{} until controlled interventions pass, and our component-rule panel shows why: taking recoverability as use licensed false causal claims in 13 of 34 noncausal cases, whereas full adjudication licensed one. A study that finds no mechanism of some kind would report a family-final negative relative to a declared basis, which later work can extend or contest by adding object classes. Reported this way, representation-level claims from different studies and about different explanations become comparable and falsifiable rather than a collection of probe accuracies and patching effects, and a systematicity benchmark can report, beside accuracy, which representational explanations of that accuracy are licensed and at what scope. To support such reporting, we release the testbed, the protocol code, and the audit reports (Appendix~\ref{app:artifact_details}).

All audits use T5 models; the verdicts have not been tested on other architectures or on large language models. The testbed is synthetic and template-clean, which makes exact interventions possible but leaves naturalistic language untested; transfer beyond our generator is shown only on one prespecified subdomain of SLOG, itself a synthetic benchmark. Both rules are least reliable at the \StatePresentOnly{} boundary, where they recover one of two and one of three held-out controls. The \Geometry{} verdict holds only within the declared closure basis and does not cover the classes excluded in Section~\ref{sec:geometry_instantiation}; its detectors remain unvalidated in trained models, and only the cross-step class has an engineered known positive. \ours{} adjudicates declared explanations and cannot discover a mechanism that no family declares, and each new family must supply and validate at least six mechanism-specific interfaces, plus a closure basis for a family-level negative. Finally, the protocol adds no new probing or intervention primitive, so its verdicts inherit the limits of the tools it adjudicates.

\begin{ack}
This work received no third-party funding. The authors declare no competing interests.
\end{ack}

\bibliographystyle{plainnat}
\bibliography{references}

\begin{thebibliography}{32}
\providecommand{\natexlab}[1]{#1}
\providecommand{\url}[1]{\texttt{#1}}
\expandafter\ifx\csname urlstyle\endcsname\relax
  \providecommand{\doi}[1]{doi: #1}\else
  \providecommand{\doi}{doi: \begingroup \urlstyle{rm}\Url}\fi

\bibitem[Alain and Bengio(2017)]{alain2016understanding}
Guillaume Alain and Yoshua Bengio.
\newblock Understanding intermediate layers using linear classifier probes.
\newblock In \emph{International Conference on Learning Representations
  (Workshop Track)}, 2017.

\bibitem[Belinkov(2022)]{belinkov2022probing}
Yonatan Belinkov.
\newblock Probing classifiers: Promises, shortcomings, and advances.
\newblock \emph{Computational Linguistics}, 48\penalty0 (1):\penalty0 207--219,
  2022.

\bibitem[Bernardi et~al.(2020)Bernardi, Benna, Rigotti, Munuera, Fusi, and
  Salzman]{bernardi2020geometry}
Silvia Bernardi, Marcus~K. Benna, Mattia Rigotti, J{\'e}r{\^o}me Munuera,
  Stefano Fusi, and C.~Daniel Salzman.
\newblock The geometry of abstraction in the hippocampus and prefrontal cortex.
\newblock \emph{Cell}, 183\penalty0 (4):\penalty0 954--967.e21, 2020.

\bibitem[Elazar et~al.(2021)Elazar, Ravfogel, Jacovi, and
  Goldberg]{elazar2021amnesic}
Yanai Elazar, Shauli Ravfogel, Alon Jacovi, and Yoav Goldberg.
\newblock Amnesic probing: Behavioral explanation with amnesic counterfactuals.
\newblock \emph{Transactions of the Association for Computational Linguistics},
  9:\penalty0 160--175, 2021.

\bibitem[Feng and Steinhardt(2024)]{feng2024binding}
Jiahai Feng and Jacob Steinhardt.
\newblock How do language models bind entities in context?
\newblock In \emph{International Conference on Learning Representations}, 2024.

\bibitem[Fodor and Pylyshyn(1988)]{fodor1988connectionism}
Jerry~A. Fodor and Zenon~W. Pylyshyn.
\newblock Connectionism and cognitive architecture: A critical analysis.
\newblock \emph{Cognition}, 28\penalty0 (1--2):\penalty0 3--71, 1988.

\bibitem[Geiger et~al.(2021)Geiger, Lu, Icard, and Potts]{geiger2021causal}
Atticus Geiger, Hanson Lu, Thomas Icard, and Christopher Potts.
\newblock Causal abstractions of neural networks.
\newblock In \emph{Advances in Neural Information Processing Systems},
  volume~34, pages 9574--9586, 2021.

\bibitem[Geiger et~al.(2024)Geiger, Wu, Potts, Icard, and
  Goodman]{geiger2024das}
Atticus Geiger, Zhengxuan Wu, Christopher Potts, Thomas Icard, and Noah
  Goodman.
\newblock Finding alignments between interpretable causal variables and
  distributed neural representations.
\newblock In \emph{Proceedings of the Third Conference on Causal Learning and
  Reasoning}, volume 236 of \emph{Proceedings of Machine Learning Research},
  pages 160--187, 2024.

\bibitem[Geirhos et~al.(2020)Geirhos, Jacobsen, Michaelis, Zemel, Brendel,
  Bethge, and Wichmann]{geirhos2020shortcut}
Robert Geirhos, J{\"o}rn-Henrik Jacobsen, Claudio Michaelis, Richard Zemel,
  Wieland Brendel, Matthias Bethge, and Felix~A. Wichmann.
\newblock Shortcut learning in deep neural networks.
\newblock \emph{Nature Machine Intelligence}, 2\penalty0 (11):\penalty0
  665--673, 2020.

\bibitem[Gupta et~al.(2024)Gupta, Arcuschin, Kwa, and
  Garriga-Alonso]{gupta2024interpbench}
Rohan Gupta, Iv{\'a}n Arcuschin, Thomas Kwa, and Adri{\`a} Garriga-Alonso.
\newblock {InterpBench}: Semi-synthetic transformers for evaluating mechanistic
  interpretability techniques.
\newblock In \emph{Advances in Neural Information Processing Systems (Datasets
  and Benchmarks Track)}, volume~37, pages 92922--92951, 2024.

\bibitem[Hewitt and Liang(2019)]{hewitt2019designing}
John Hewitt and Percy Liang.
\newblock Designing and interpreting probes with control tasks.
\newblock In \emph{Proceedings of the 2019 Conference on Empirical Methods in
  Natural Language Processing and the 9th International Joint Conference on
  Natural Language Processing ({EMNLP}-{IJCNLP})}, pages 2733--2743.
  Association for Computational Linguistics, 2019.

\bibitem[Ito et~al.(2022)Ito, Klinger, Schultz, Murray, Cole, and
  Rigotti]{ito2022compositional}
Takuya Ito, Tim Klinger, Doug Schultz, John Murray, Michael Cole, and Mattia
  Rigotti.
\newblock Compositional generalization through abstract representations in
  human and artificial neural networks.
\newblock In \emph{Advances in Neural Information Processing Systems},
  volume~35, pages 32225--32239, 2022.

\bibitem[Jacovi and Goldberg(2020)]{jacovi2020faithfulness}
Alon Jacovi and Yoav Goldberg.
\newblock Towards faithfully interpretable {NLP} systems: How should we define
  and evaluate faithfulness?
\newblock In \emph{Proceedings of the 58th Annual Meeting of the Association
  for Computational Linguistics}, pages 4198--4205. Association for
  Computational Linguistics, 2020.

\bibitem[Keysers et~al.(2020)Keysers, Sch{\"a}rli, Scales, Buisman, Furrer,
  Kashubin, Momchev, Sinopalnikov, Stafiniak, Tihon, Tsarkov, Wang, van Zee,
  and Bousquet]{keysers2020measuring}
Daniel Keysers, Nathanael Sch{\"a}rli, Nathan Scales, Hylke Buisman, Daniel
  Furrer, Sergii Kashubin, Nikola Momchev, Danila Sinopalnikov, Lukasz
  Stafiniak, Tibor Tihon, Dmitry Tsarkov, Xiao Wang, Marc van Zee, and Olivier
  Bousquet.
\newblock Measuring compositional generalization: A comprehensive method on
  realistic data.
\newblock In \emph{International Conference on Learning Representations}, 2020.

\bibitem[Kim and Linzen(2020)]{kim2020cogs}
Najoung Kim and Tal Linzen.
\newblock {COGS}: A compositional generalization challenge based on semantic
  interpretation.
\newblock In \emph{Proceedings of the 2020 Conference on Empirical Methods in
  Natural Language Processing ({EMNLP})}, pages 9087--9105. Association for
  Computational Linguistics, 2020.

\bibitem[Lake and Baroni(2018)]{lake2018generalization}
Brenden Lake and Marco Baroni.
\newblock Generalization without systematicity: On the compositional skills of
  sequence-to-sequence recurrent networks.
\newblock In \emph{Proceedings of the 35th International Conference on Machine
  Learning}, volume~80 of \emph{Proceedings of Machine Learning Research},
  pages 2873--2882, 2018.

\bibitem[Lake and Baroni(2023)]{lake2023human}
Brenden~M. Lake and Marco Baroni.
\newblock Human-like systematic generalization through a meta-learning neural
  network.
\newblock \emph{Nature}, 623\penalty0 (7985):\penalty0 115--121, 2023.

\bibitem[Li et~al.(2023)Li, Donatelli, Koller, Linzen, Yao, and
  Kim]{li2023slog}
Bingzhi Li, Lucia Donatelli, Alexander Koller, Tal Linzen, Yuekun Yao, and
  Najoung Kim.
\newblock {SLOG}: A structural generalization benchmark for semantic parsing.
\newblock In \emph{Proceedings of the 2023 Conference on Empirical Methods in
  Natural Language Processing}, pages 3213--3232. Association for Computational
  Linguistics, 2023.

\bibitem[Makelov et~al.(2024)Makelov, Lange, Geiger, and
  Nanda]{makelov2024subspace}
Aleksandar Makelov, Georg Lange, Atticus Geiger, and Neel Nanda.
\newblock Is this the subspace you are looking for? {An} interpretability
  illusion for subspace activation patching.
\newblock In \emph{International Conference on Learning Representations}, 2024.

\bibitem[McCoy et~al.(2019{\natexlab{a}})McCoy, Linzen, Dunbar, and
  Smolensky]{mccoy2019rnns}
R.~Thomas McCoy, Tal Linzen, Ewan Dunbar, and Paul Smolensky.
\newblock {RNNs} implicitly implement tensor-product representations.
\newblock In \emph{International Conference on Learning Representations},
  2019{\natexlab{a}}.

\bibitem[McCoy et~al.(2019{\natexlab{b}})McCoy, Pavlick, and
  Linzen]{mccoy2019right}
R.~Thomas McCoy, Ellie Pavlick, and Tal Linzen.
\newblock Right for the wrong reasons: Diagnosing syntactic heuristics in
  natural language inference.
\newblock In \emph{Proceedings of the 57th Annual Meeting of the Association
  for Computational Linguistics}, pages 3428--3448. Association for
  Computational Linguistics, 2019{\natexlab{b}}.

\bibitem[Meng et~al.(2022)Meng, Bau, Andonian, and Belinkov]{meng2022locating}
Kevin Meng, David Bau, Alex Andonian, and Yonatan Belinkov.
\newblock Locating and editing factual associations in {GPT}.
\newblock In \emph{Advances in Neural Information Processing Systems},
  volume~35, pages 17359--17372, 2022.

\bibitem[Mueller et~al.(2025)Mueller, Geiger, Wiegreffe, Arad, Arcuschin,
  Belfki, Chan, Fiotto-Kaufman, Haklay, Hanna, Huang, Gupta, Nikankin, Orgad,
  Prakash, Reusch, Sankaranarayanan, Shao, Stolfo, Tutek, Zur, Bau, and
  Belinkov]{mueller2025mib}
Aaron Mueller, Atticus Geiger, Sarah Wiegreffe, Dana Arad, Iv{\'a}n Arcuschin,
  Adam Belfki, Yik~Siu Chan, Jaden~Fried Fiotto-Kaufman, Tal Haklay, Michael
  Hanna, Jing Huang, Rohan Gupta, Yaniv Nikankin, Hadas Orgad, Nikhil Prakash,
  Anja Reusch, Aruna Sankaranarayanan, Shun Shao, Alessandro Stolfo, Martin
  Tutek, Amir Zur, David Bau, and Yonatan Belinkov.
\newblock {MIB}: A mechanistic interpretability benchmark.
\newblock In \emph{Proceedings of the 42nd International Conference on Machine
  Learning}, volume 267 of \emph{Proceedings of Machine Learning Research},
  pages 45069--45108, 2025.

\bibitem[Raffel et~al.(2020)Raffel, Shazeer, Roberts, Lee, Narang, Matena,
  Zhou, Li, and Liu]{raffel2020t5}
Colin Raffel, Noam Shazeer, Adam Roberts, Katherine Lee, Sharan Narang, Michael
  Matena, Yanqi Zhou, Wei Li, and Peter~J. Liu.
\newblock Exploring the limits of transfer learning with a unified text-to-text
  transformer.
\newblock \emph{Journal of Machine Learning Research}, 21\penalty0
  (140):\penalty0 1--67, 2020.

\bibitem[Ravichander et~al.(2021)Ravichander, Belinkov, and
  Hovy]{ravichander2020probing}
Abhilasha Ravichander, Yonatan Belinkov, and Eduard Hovy.
\newblock Probing the probing paradigm: Does probing accuracy entail task
  relevance?
\newblock In \emph{Proceedings of the 16th Conference of the European Chapter
  of the Association for Computational Linguistics: Main Volume}, pages
  3363--3377. Association for Computational Linguistics, 2021.

\bibitem[Smolensky(1990)]{smolensky1990tensor}
Paul Smolensky.
\newblock Tensor product variable binding and the representation of symbolic
  structures in connectionist systems.
\newblock \emph{Artificial Intelligence}, 46\penalty0 (1--2):\penalty0
  159--216, 1990.

\bibitem[Soulos et~al.(2020)Soulos, McCoy, Linzen, and
  Smolensky]{soulos2020discovering}
Paul Soulos, R.~Thomas McCoy, Tal Linzen, and Paul Smolensky.
\newblock Discovering the compositional structure of vector representations
  with role learning networks.
\newblock In \emph{Proceedings of the Third BlackboxNLP Workshop on Analyzing
  and Interpreting Neural Networks for NLP}, pages 238--254. Association for
  Computational Linguistics, 2020.

\bibitem[Vegner et~al.(2025)Vegner, de~Souza, Forch, Lewis, and
  Doumas]{vegner2025behavioural}
Ivan Vegner, Sydelle de~Souza, Valentin Forch, Martha Lewis, and Leonidas A.~A.
  Doumas.
\newblock Behavioural vs. representational systematicity in end-to-end models:
  An opinionated survey.
\newblock In \emph{Proceedings of the 63rd Annual Meeting of the Association
  for Computational Linguistics (Volume 1: Long Papers)}, pages 31842--31856.
  Association for Computational Linguistics, 2025.

\bibitem[Vig et~al.(2020)Vig, Gehrmann, Belinkov, Qian, Nevo, Singer, and
  Shieber]{vig2020investigating}
Jesse Vig, Sebastian Gehrmann, Yonatan Belinkov, Sharon Qian, Daniel Nevo,
  Yaron Singer, and Stuart Shieber.
\newblock Investigating gender bias in language models using causal mediation
  analysis.
\newblock In \emph{Advances in Neural Information Processing Systems},
  volume~33, pages 12388--12401, 2020.

\bibitem[Wang et~al.(2023)Wang, Variengien, Conmy, Shlegeris, and
  Steinhardt]{wang2023interpretability}
Kevin~Ro Wang, Alexandre Variengien, Arthur Conmy, Buck Shlegeris, and Jacob
  Steinhardt.
\newblock Interpretability in the wild: {A} circuit for indirect object
  identification in {GPT}-2 small.
\newblock In \emph{International Conference on Learning Representations}, 2023.

\bibitem[Wu et~al.(2023)Wu, Manning, and Potts]{wu2023recogs}
Zhengxuan Wu, Christopher~D. Manning, and Christopher Potts.
\newblock {ReCOGS}: How incidental details of a logical form overshadow an
  evaluation of semantic interpretation.
\newblock \emph{Transactions of the Association for Computational Linguistics},
  11:\penalty0 1719--1733, 2023.

\bibitem[Zhang and Nanda(2024)]{zhang2024patching}
Fred Zhang and Neel Nanda.
\newblock Towards best practices of activation patching in language models:
  Metrics and methods.
\newblock In \emph{International Conference on Learning Representations}, 2024.

\end{thebibliography}

\appendix

\FloatBarrier
\section{Data Construction Details}
\label{app:data}

Each matched set is created from one semantic design: a predicate, an agent filler, a theme filler, alternate fillers for the replacement items, an agent modifier, a theme modifier, a voice, and an optional relative-clause context. The matched easy source uses an active, non-embedded realization, and the hard target keeps the gold meaning in the set's construction, so in the 28 active sets without embedding the two coincide. The other five items each change one contrast: agent replacement, theme replacement, role swap, filler replacement, and modifier swap. The gold output is a logical form that names the predicate, its role fillers, and their modifiers.

The renderer produces three surface candidates per item, and validation checks that the required fillers, modifiers, and predicate surface are present, that the active or passive order is consistent, that the relative-clause context appears when required, and that no pronouns occur. All 2{,}016 candidates are valid, and one realization is selected per item, which gives 672 canonical items in 96 complete matched sets with exact counterpart coverage 1.0, 192 interchange pairs, and 96 repair pairs. The sets comprise 28 active and 31 passive sets without embedding and 31 active and 6 passive sets with a relative clause (196, 217, 217, and 42 items). The generator uses seed 13.

Splits are drawn at the level of matched sets. The 64 training sets (22 active and 21 passive sets without embedding and 21 active sets with a relative clause, with 154, 147, and 147 items) contain 448 items and no passive item with a relative clause. Of the 32 held-out sets, 16 form the development split with all seven items, and 16 form the generalization split, whose six non-source items are scored for generalization while their matched easy sources join the development split; this gives 128 development and 96 generalization items, and all six passive sets with a relative clause (42 items) are held out. A surface-shifted realization that keeps the longest valid template per item, built for 48 of the 96 matched sets (224 training items), supplies the surface-shifted targets. Earlier, less controlled surface variants are retained in the artifact as stress tests but enter no reported decision, because the intervention protocol requires exact counterpart coverage.

\FloatBarrier
\section{Detailed Evidence Blocks}
\label{app:evidence_blocks}

Section~\ref{sec:evidence} summarizes the evidence layer by role, and Table~\ref{tab:app_evidence_blocks} gives the block-level specification. Each block produces metrics that enter the behavioral anchor, the presence score, the causal-use score, or the validity gates, and no single block can license \StateCausalUse{}.

\begin{table}[ht]
\centering
\small
\begin{tabular}{@{}p{0.16\linewidth}p{0.25\linewidth}p{0.25\linewidth}p{0.24\linewidth}@{}}
\toprule
Block & Core question & Evidence produced & Controls or rejection checks \\
\midrule
Behavioral anchor & Does the model exhibit the target systematic behavior on the matched contrast? & Exact match and clause-level F1 between generated and gold logical forms; pass, fail, and control strata. & Used only as an anchor; cannot by itself assign a state or produce \StateCausalUse{}. \\
Presence readout & Is the family-specific target signal recoverable from hidden states? & Readout accuracy, controlled readout gaps, and stability across matched items. & Surface position, filler identity, predicate identity, and other simple correlates. \\
Interchange intervention & Does transferring the signal across matched items change generation in the predicted direction? & Intervention effect and advantage over the strongest matched control. & Wrong-role, non-role, and shuffled controls. \\
Necessity intervention & Does removing or dampening the signal selectively degrade the target output? & Selective degradation of target parses relative to controls. & Degradation must be selective rather than a global disruption of generation. \\
Repair intervention & Does inserting a matched source signal into a hard target repair its output? & Repair advantage on hard targets over controlled repair baselines. & Path-only, wrong-role, non-role, shuffled, and generic-decoder-gain controls. \\
Item linkage & Do improvements occur on the items that the explanation predicts? & Item-linked effect pass rate and aggregate-versus-item diagnostics. & Rejects aggregate-only improvements not linked to intended items. \\
Realism gate & Do interventions stay within the model's natural computational regime? & Natural-state realism pass rate of intervention states. & Rejects off-manifold or unrealistic interventions. \\
\bottomrule
\end{tabular}
\caption{Evidence blocks of \ours{}. The main text groups them into behavioral anchoring, recoverability, intervention tests, and validity checks.}
\label{tab:app_evidence_blocks}
\end{table}

The behavioral anchor computes exact match and clause-level F1, which gives partial credit for correctly generated semantic atoms and defines behavioral pass, fail, and control strata. Presence readouts provide recoverability evidence only; for \Binding{}, the role readout is the leave-one-out nearest-centroid accuracy of semantic roles on final-layer hidden states, and the surface-control gap compares it with a readout of surface position. A high readout means that a signal can be recovered, not that the model uses it.

The intervention blocks provide causal-use evidence only when their effects are selective and control-calibrated. Interchange compares a matched signal transfer with wrong-role, non-role, and shuffled alternatives; necessity tests whether removing the signal causes selective semantic degradation rather than generic output damage; and repair tests whether inserting a matched source signal improves the hard target beyond path-only, wrong-role, non-role, shuffled, and generic decoder-gain controls. Realism and item linkage are gates rather than score features: they block stronger states when intervention effects are off-manifold or not tied to the intended items.

\FloatBarrier
\section{Full Adjudicator Configuration}
\label{app:adjudicator_config}

Section~\ref{sec:scoring} describes the decision layer, and this appendix records the configuration used for the reported audits. The tables below list the predeclared features, boundaries, scales, weights, gates, and calibration sources that can affect a state decision; all of them are frozen on calibration controls before target and closure outcomes are interpreted.

\subsection{Score features}
\label{app:score_features}

Table~\ref{tab:app_score_features} lists the score features, with their evidence group, scoring configuration, and the calibration source that fixed the boundary.

\begin{table}[ht]
\centering
\small
\begin{tabular}{@{}p{0.16\linewidth}p{0.29\linewidth}p{0.29\linewidth}p{0.16\linewidth}@{}}
\toprule
Evidence group & Feature & Scoring configuration & Calibration source \\
\midrule
Behavior & Clause-F1 anchor & Direction $+$; boundary $\tau=0.570312$ with floor $0.070312$; scale $s=0.10$; weight $0.15$ prior. & Behavioral pass/fail controls \\
Presence & Semantic-role readout accuracy & Direction $+$; boundary $\tau=0.914062$; scale $s=0.05$; weight $0.60$. & State-validating controls \\
Presence & Surface-control gap & Direction $+$; boundary $\tau=-0.085938$; scale $s=0.05$; weight $0.40$. & Surface, filler, and predicate controls \\
Use & Interchange or repair advantage over strongest control & Direction $+$; boundary $\tau=0.245002$; scale $s=0.05$; weight $0.60$. & \StateCausalUse{} vs.\ \StatePresentOnly{} controls \\
Use & Permutation or shuffled-control significance & Direction $-$; boundary $\tau=0.05$; scale $s=0.01$; weight $0.40$. & Shuffled and wrong-role controls \\
\bottomrule
\end{tabular}
\caption{Score features of the reported audits. Weights are normalized within each group except for the behavior prior. Gates and required-status constraints are given in Table~\ref{tab:app_gate_config}.}
\label{tab:app_score_features}
\end{table}

\subsection{Gates and stop rules}
\label{app:gates_stop_rules}

Table~\ref{tab:app_gate_config} lists the gates and stop rules. Unlike score features, they are not averaged into a group score; they are required conditions for licensing the corresponding state.

\begin{table}[ht]
\centering
\small
\begin{tabular}{@{}p{0.19\linewidth}p{0.31\linewidth}p{0.19\linewidth}p{0.21\linewidth}@{}}
\toprule
Gate & Criterion & Required for & Calibration source \\
\midrule
Realism gate & Natural-state realism pass rate must exceed $0.60$. & Intervention states & Realism calibration controls \\
Linkage gate & Item-linked effect pass rate must exceed $0.50$. & Intervention states & Item-linkage controls \\
Coverage gate & Required evidence coverage must exceed $0.80$. & All state decisions & Held-out control panels \\
Decision gate & Top-state score margin must exceed $0.10$. & States stronger than \StateUndetermined{} & Held-out control panels \\
\bottomrule
\end{tabular}
\caption{Gates and stop rules. Gate rows are required conditions rather than score features.}
\label{tab:app_gate_config}
\end{table}

The behavior feature acts only as a low-weight prior and cannot by itself produce \StateCausalUse{}. Presence features support recoverability when the target signal is readable beyond surface and identity controls. Use features support causal use only when the intervention or repair effect exceeds the strongest matched control and the required validity gates pass. The coverage and decision-margin gates prevent stronger states when evidence is incomplete or the top state is not sufficiently separated from the alternatives.

\FloatBarrier
\section{End-to-End Walkthrough}
\label{app:walkthrough}

This appendix follows one matched set, MS-047, through the three layers of a \Binding{} audit of a normally fine-tuned T5-small target. The set instantiates the seven-item construction of Appendix~\ref{app:data} from an explicit semantic design, and the panel-level behavioral anchor that licenses the audit is evaluated separately and passes. Table~\ref{tab:walkthrough} lists the evidence at each stage.

\begin{table}[ht]
\centering
\small
\begin{tabular}{@{}p{0.17\linewidth}p{0.47\linewidth}p{0.28\linewidth}@{}}
\toprule
Stage & Evidence & Result \\
\midrule
Behavior & Exact match 6/7 $=0.857$; clause-F1 0.947. & Behavior score 0.977. \\
Presence & Role readout 0.938; surface-control gap 0.071. & Both presence features pass. \\
Controlled use & Interchange 0.512 vs.\ strongest control 0.181; necessity 0.276 vs.\ global control 0.058; repair 0.464 vs.\ strongest control 0.161. & Interchange and repair advantages 0.331 and 0.303. \\
Validity and decision & Realism 0.714, linkage 0.571, coverage 0.857. & All gates pass; final state \StateCausalUse{}. \\
\bottomrule
\end{tabular}
\caption{End-to-end trace of matched set MS-047 in the \Binding{} audit.}
\label{tab:walkthrough}
\end{table}

Each number maps onto the configuration of Appendix~\ref{app:adjudicator_config}. The behavior score applies Eq.~\eqref{eq:score_transform} to clause-F1 with $\tau=0.570312$ and $s=0.10$, so $\sigmoid((0.947-0.570)/0.10)=0.977$; behavior enters only as a prior, so this score cannot decide the state. The role readout and the surface-control gap exceed their boundaries of 0.914 and $-0.086$, so presence is supported. The interchange and repair advantages exceed the use boundary of 0.245, and necessity exceeds its global control. Realism ($0.714>0.60$), linkage ($0.571>0.50$), and coverage ($0.857>0.80$) pass their gates, and the set is \StateCausalUse{}. The released repository contains the seven inputs of MS-047 with their gold logical forms and exact pairs, and a script that runs this trace on a checkpoint with the same audit and decision function (Appendix~\ref{app:artifact_details}).

\FloatBarrier
\section{Reusable and Family-Specific Scaffold}
\label{app:scaffold_components}

The shared scaffold fixes the state semantics, evidence slots, scoring grammar, gates, stop rules, closure driver, and reporting format, whereas each explanatory family specifies the signal it claims to explain, the operations that detect and manipulate that signal, and the controls that would count against the explanation (Table~\ref{tab:app_scaffold_components}). This division is the extension contract of the artifact: a new family reuses the matched-set format, state names, evidence slots, and report structure, and supplies its own operationalization of the hypothesized signal, which keeps each state tied to a declared explanation.

\begin{table}[ht]
\centering
\small
\begin{tabular}{@{}p{0.19\linewidth}p{0.34\linewidth}p{0.34\linewidth}@{}}
\toprule
Layer & Shared component & What a new explanatory family must provide \\
\midrule
Data & Matched sets for the behavioral contrast. & Usually unchanged when the family explains the same contrast; redesigned when the contrast changes. \\
Evidence slots & Behavioral anchor, presence readout, interchange, necessity, repair, item linkage, and gates. & Target signal, detector, interchange, necessity, and repair operations, and controls. \\
Adjudication & State semantics, scoring grammar, gates, stop rules, and closure driver. & Control-fitted feature boundaries and, for a family-level negative, a closure basis. \\
Artifact & Data generator, protocol runners, reports, and documentation. & Documentation of operations, expected runtime, and failure modes. \\
\bottomrule
\end{tabular}
\caption{Reusable and family-specific parts of \ours{}.}
\label{tab:app_scaffold_components}
\end{table}

The extension cost is measurable. In our implementation, the shared schema, scoring, gates and stop rules, closure driver, and reporting span 18 files and 2{,}164 lines of code, and the \Binding{} and \Geometry{} operations span another 18 files and 3{,}584 lines. Porting both existing families to T5-base touched 257 lines and took 11.4 development hours with no interface change, and porting the \Binding{} chain to the SLOG subdomain required a 705-line adapter with no change to the family operations. These are ports of existing families. A new family must instantiate and validate at least six mechanism-specific interfaces, namely its target signal, detector, interchange, necessity, and repair operations, and controls, and a seventh, a closure basis, for a family-level negative.

\FloatBarrier
\section{Per-Family Operationalization Details}
\label{app:families}

For \Binding{}, the object is a role--filler signal. Interchange patches a role-indexed hidden vector or an aggregated role signal from a matched source into a matched recipient, necessity removes or dampens role-relevant information, and repair moves the corresponding role--filler signal from the matched easy source into the hard target. \Binding{} calibration controls are T5-small checkpoints trained on the full training split (\StateCausalUse{}), the same checkpoints with decoder cross-attention scaled to 0.25 (\StatePresentOnly{}), and checkpoints trained on permuted targets or on 64 or 128 items (\StateUnsupported{}). Its target panel contains the normally fine-tuned target, two checkpoints with decoder cross-attention scaled to 0.25 (one raw-prompted and one instruction-prompted), a target trained on 128 items, and a target trained on the surface-shifted realization.

For \Geometry{}, the object is not assumed to be a single vector. The operations include subspace projection and removal, relation-subspace tests, trajectory and layer-path summaries, cross-attention access and bottlenecking, cross-step bridges, and query--source relations; they correspond functionally to the \Binding{} interventions, but manipulate geometric relations, routes, or trajectories. \Geometry{} has an independent control suite of T5-small checkpoints trained with a fixed encoder feature mask (keep ratio 0.75): fully trained checkpoints serve as \StateCausalUse{} controls; checkpoints whose decoder access is weakened, by scaling cross-attention to 0.25 at inference or during training or by moderately masking the decoder readout, serve as \StatePresentOnly{} controls; and checkpoints trained on permuted targets or on a small subset of the training split, or with a strongly masked readout, serve as \StateUnsupported{} controls. Variants A--D were built to produce \Geometry{} \StatePresentOnly{} regimes: Variant~A blocks cross-attention access, promoted Variant~A applies deterministic layer policies to it, Variant~B projects out a decoder relation subspace, Variant~C imposes a low-rank cross-attention bottleneck, and Variant~D attenuates a seeded relation-bearing subspace of the cross-attention output.

\subsection{Geometry closure basis}
\label{app:closure_basis}

An object class enters the closure basis only if it supports a complete chain from an observable object and a matched detector to object-specific intervention and repair and falsifying controls. Table~\ref{tab:app_closure_basis} lists the three classes that meet this criterion. Layer trajectories and routes are kept as diagnostics because the artifact lacks a closure-grade repair and control chain for them. Nonlinear manifolds, higher-order or polysemantic interactions, sparse-feature or sparse-autoencoder circuits, diffuse long-range dynamic routes, and nonlocal multi-source mixtures lie outside the basis; they are omitted classes, not negative cases within it.

\begin{table}[ht]
\centering
\small
\begin{tabular}{@{}p{0.20\linewidth}p{0.27\linewidth}p{0.45\linewidth}@{}}
\toprule
Object class & Matched detector & What the detector tests \\
\midrule
Interaction objects & Predicate-conditioned relation detector & A layer-local relation subspace read conditionally on the predicate. \\
Cross-step bridges & Cross-step bridge detector (cross-step alignment) & Whether relation information forms a traceable route across decoder steps; a max-statistic controls multiplicity over the frozen candidate search. \\
Query--source relations & Argument-locus detector & Whether a decoding query bears a structured relation to the relevant source argument. \\
\bottomrule
\end{tabular}
\caption{Frozen \Geometry{} closure basis and matched detectors.}
\label{tab:app_closure_basis}
\end{table}

Each class has 12 candidate objects, formed from its decoding steps and templates that vary the head choice (top-1 head or a top-2 bundle), the basis (single or bundled), and the storage variant (matched store or none). Before closure, three step aligners (leftmost exact subsequence, slot-aware literal field, and teacher-forced bridge) were compared on the frozen controls; all align predicate steps at rate 1.0 and argument steps at rate 0.0, so observability is sealed as predicate-dominant with aligner A1. A candidate survives discovery if its contrast on the calibration controls exceeds a minimum gap, and a class is accepted when at least 8 of its 12 candidates survive, covering at least two decoding steps or step pairs and two templates. Closure then requires, for a family-final \StateUnsupported{}, that observability be sealed, that every class finish discovery and no object pass the causal-use (necessity) gate, with generic decoder gain as a mandatory matched control, that the direct and slice-local \StatePresentOnly{} routes fail on the decoder-scaled, trained-scaled, and readout-masked \StatePresentOnly{} regimes, and that the \StateUnsupported{} and \StateCausalUse{} controls remain stable.

\FloatBarrier
\section{Additional Result Details}
\label{app:result_details}

This appendix reports the per-panel details behind Section~\ref{sec:results}, followed by the robustness, truth-control, and transfer analyses that test the two headline verdicts and by the comparison of full adjudication with its component rules. Table~\ref{tab:app_checks} indexes these analyses.

\begin{table}[ht]
\centering
\small
\begin{tabular}{@{}p{0.18\linewidth}p{0.27\linewidth}p{0.47\linewidth}@{}}
\toprule
Analysis & Design & Conclusion \\
\midrule
Parameter sensitivity & 729-configuration grid; 487 calibration-compatible & Both verdicts hold in all 487 configurations; the nearest closure flip lies outside the compatible region. \\
Training seeds & Five T5-small seeds & \Binding{} \StateCausalUse{} and \Geometry{} \StateUnsupported{} in 5/5. \\
Matched-set bootstrap & 1{,}000 cluster replicates over 96 sets & No \Geometry{} object passes closure; no \StateCausalUse{} reversal. \\
Truth controls & Engineered triplets, three seeds & The cross-step \Geometry{} branch and \Binding{} recover all three evidential states. \\
Component rules & 41 predeclared cases with known causal status & Full adjudication licenses the fewest false \StateCausalUse{} states and keeps all causal cases. \\
T5-base transfer & Three seeds; frozen and recalibrated rules & Both verdicts replicate in 6/6 seed-by-rule runs. \\
SLOG transfer & 96-pair subdomain; three seeds & The \Binding{} chain returns \StateCausalUse{} in two seeds and \StateUndetermined{} in one. \\
\bottomrule
\end{tabular}
\caption{Analyses that test the two headline verdicts, and the comparison of full adjudication with its component rules.}
\label{tab:app_checks}
\end{table}

\subsection{Binding calibration and target details}
\label{app:binding_result_details}

\begin{table}[ht]
\centering
\small
\begin{tabular}{@{}p{0.22\linewidth}p{0.11\linewidth}p{0.28\linewidth}p{0.29\linewidth}@{}}
\toprule
Panel & $n$ & Outcome & Note \\
\midrule
Initial calibration & 3 & all intended states & Separates \StateUnsupported{}, \StatePresentOnly{}, and \StateCausalUse{} controls. \\
Held-out controls & 6 & acc.\ 0.833; macro-F1 0.822 & Two controls per state. \\
Behavior-feature ablation & 6 & no change in acc.\ or macro-F1 & Behavior feature share is 0.041. \\
Initial target audit & 3 & all \StateCausalUse{} & Mean top-state score and decision margin are both 1.0. \\
Heterogeneous target audit & 5 & 1 \StateCausalUse{}, 1 \StatePresentOnly{}, 1 \StateUnsupported{}, 2 \StateUndetermined{} & Normally trained, weak, decoder-scaled, instruction-prompted scaled, and surface-shifted targets. \\
\bottomrule
\end{tabular}
\caption{\Binding{} calibration and target-panel details.}
\label{tab:app_binding_details}
\end{table}

Table~\ref{tab:app_binding_details} shows that the held-out \Binding{} panel separates the \StateUnsupported{} and \StateCausalUse{} regimes and that its one error escalates a \StatePresentOnly{} control to \StateCausalUse{}. The calibration therefore checks the state semantics; it is not a claim of probabilistic calibration.

\subsection{Geometry candidate and variant details}
\label{app:geometry_variant_details}

\begin{table}[ht]
\centering
\small
\begin{tabular}{@{}p{0.22\linewidth}p{0.08\linewidth}p{0.13\linewidth}p{0.13\linewidth}p{0.13\linewidth}p{0.14\linewidth}@{}}
\toprule
\Geometry{} panel & $n$ & \StateCausalUse{} & \StatePresentOnly{} & \StateUnsupported{} & \StateUndetermined{} \\
\midrule
Pre-closure target audit & 7 & 2 & 1 & 3 & 1 \\
Variant A & 6 & 3 & 1 & 2 & 0 \\
Promoted Variant A & 6 & 0 & 1 & 5 & 0 \\
Variant B & 6 & 0 & 0 & 6 & 0 \\
Variant C & 6 & 0 & 0 & 6 & 0 \\
Variant D & 6 & 6 & 0 & 0 & 0 \\
\bottomrule
\end{tabular}
\caption{\Geometry{} pre-closure target and variant outcomes. These are operation-level outcomes and do not by themselves license a family-level \Geometry{} claim.}
\label{tab:app_geometry_variants}
\end{table}

The nine held-out \Geometry{} controls comprise two \StateCausalUse{}, three \StatePresentOnly{}, and four \StateUnsupported{} regimes. The frozen rule recalls \StateUnsupported{} and \StateCausalUse{} controls perfectly but \StatePresentOnly{} controls at 0.333, and returns \StateUndetermined{} once (accuracy 0.778, macro-F1 0.796). On the target panel (Table~\ref{tab:app_geometry_variants}), the two fully trained targets are \StateCausalUse{}; the weak target, a trained intermediate regime, and a strongly readout-masked regime are \StateUnsupported{}; a second trained intermediate regime is \StatePresentOnly{}; and a moderately readout-masked regime is \StateUndetermined{}. The variants either collapse to one state or retain a single \StatePresentOnly{} case. Variant~D, which is \StateCausalUse{} in all six runs, shows why these outcomes stay at the operation level: its interchange effect is strongly positive against all controls, but its necessity effect is tied with the strongest control and repair evidence is unavailable, so it supplies no object-specific necessity of the kind closure requires.

\subsection{Detector-truthfulness details}
\label{app:detector_truthfulness_details}

\begin{table}[ht]
\centering
\small
\begin{tabular}{@{}p{0.28\linewidth}p{0.18\linewidth}p{0.18\linewidth}p{0.24\linewidth}@{}}
\toprule
Detector class & Canonical examples & Coverage & Validation outcome \\
\midrule
Argument locus & 336 & 1.0 & not validated \\
Predicate-conditioned relation & 336 & 1.0 & not validated \\
Cross-step bridge & 336 & 1.0 & not validated \\
\bottomrule
\end{tabular}
\caption{Detector-truthfulness checks for geometry-like objects.}
\label{tab:app_detector_truthfulness}
\end{table}

Detector truthfulness is separate from closure. Closure asks whether any admissible object in the frozen basis survives discovery and passes the required gates; detector truthfulness asks whether the detectors themselves monitor geometry-like objects correctly under known positive, negative, and null conditions in trained models. Each detector class is tested on 336 canonical examples with full coverage in two stages, with 8 and 6 targets. All three classes are anchorable, but none passes the full validation: the argument-locus and cross-step detectors fail detector-specific observability, and all three fail the necessity and \StatePresentOnly{} tests on known positives (Table~\ref{tab:app_detector_truthfulness}). The strongest licensed \Geometry{} conclusion therefore remains family-final.

\subsection{Sensitivity, training seeds, and matched-set bootstrap}
\label{app:robustness}

The sensitivity analysis varies every verdict-relevant parameter block over a predeclared grid of six blocks with three levels each ($3^6=729$ configurations), spanning score boundaries, scales, and weights, the realism, linkage, coverage, and margin gates, and the \Geometry{} discovery, necessity, and null thresholds. A configuration is calibration-compatible if it preserves the intended states of the calibration controls, a rule that does not look at any target; 487 configurations are compatible. Across all 487, the claim-bearing \Binding{} target remains \StateCausalUse{}, and no \Geometry{} object passes closure, so the family keeps its family-final \StateUnsupported{} state. A joint worst-case search that favors \Geometry{} and disfavors \Binding{} produces no change in either verdict, and the nearest \Geometry{} closure flip lies outside the compatible region at normalized distance 3.75. The only rows that change are peripheral and candidate rows that carry no claim, including \StatePresentOnly{}/\StateUndetermined{} transitions and one candidate \StateCausalUse{}/\StateUndetermined{} transition.

Across five fixed, independent T5-small training seeds, \Binding{} is \StateCausalUse{} in 5/5, and no \Geometry{} object passes closure, leaving the family \StateUnsupported{}, in 5/5. For seed 0, a cluster bootstrap that resamples the 96 matched sets 1{,}000 times yields no \Geometry{} object passing closure in any replicate, with a family-final \StateUnsupported{} state in 998 and \StateUndetermined{} in 2, and never \StateCausalUse{}. The 95\% bootstrap intervals for the \Binding{} evidence are 0.337 [0.301, 0.371] for interchange, 0.284 [0.249, 0.321] for necessity, 0.318 [0.279, 0.356] for repair, 0.729 [0.667, 0.786] for realism, and 0.646 [0.583, 0.708] for linkage, all on the supporting side of their decision boundaries.

\subsection{Truth controls}
\label{app:truth_controls}

The truth controls test whether each pipeline can return all three evidential states when the truth is known by construction. For the cross-step \Geometry{} class and for \Binding{}, matched triplets with the same architecture, task, budget, and paired seeds place a designated signal in one of three conditions: present and causally required, equally recoverable but not causally used, or replaced by matched null content. With blind discovery and independent confirmation, the \Geometry{} pipeline recovers the designated object in 3/3 causal and 3/3 \StatePresentOnly{} seeds. Both pipelines return \StateCausalUse{}, \StatePresentOnly{}, and \StateUnsupported{} correctly in 3/3 seeds each, and neither produces a false \StateCausalUse{} in its six noncausal runs. In the causal condition, the \Geometry{} correct-object advantage is 0.335 against 0.309 for \Binding{}, a difference of 0.026 within the pre-frozen $\pm0.060$ band, and above the use boundary of 0.245002. These controls validate engineered in-basis sensitivity for the cross-step class only. \Binding{} does not carry the same blind-localization and multiplicity burden, and no equivalence interval was estimated, so the comparison does not establish equal power across families.

\subsection{Component-rule comparison}
\label{app:component_rules}

The component-rule comparison asks whether full adjudication changes conclusions relative to its components. A frozen panel contains 41 predeclared cases whose causal status is known by construction, 7 causal and 34 noncausal. Three license rules are applied to the same evidence (Table~\ref{tab:app_component_rules}): the heuristic that recoverability implies causal use, a rule that licenses causal use from intervention effects without the validity gates and stop rules, and full adjudication.

\begin{table}[ht]
\centering
\small
\begin{tabular}{@{}p{0.34\linewidth}p{0.22\linewidth}p{0.17\linewidth}p{0.17\linewidth}@{}}
\toprule
License rule & False \StateCausalUse{} (noncausal cases) & Causal cases retained & Abstentions \\
\midrule
Recoverable $\Rightarrow$ causal & 13/34 & 7/7 & -- \\
Ungated intervention & 7/34 & 7/7 & -- \\
Full \ours{} & 1/34 & 7/7 & 6/41 \\
\bottomrule
\end{tabular}
\caption{False \StateCausalUse{} licenses on 41 cases with known causal status. Full adjudication covers 35 of the 41 cases.}
\label{tab:app_component_rules}
\end{table}

Taking recoverability as evidence of use licenses \StateCausalUse{} in 13 of the 34 noncausal cases, and licensing it from ungated interventions does so in 7. Full adjudication does so in 1, retains all 7 causal cases, and abstains on 6 cases.

\subsection{T5-base and SLOG transfer}
\label{app:transfer}

The T5-base audit uses three fixed training seeds and freezes the testbed, split, state semantics, family definitions, closure classes, controls, reporting schema, and target-independent relative-depth, head, and rank mappings. It applies both the T5-small rules directly and rules recalibrated only on T5-base calibration controls. All three behavioral anchors pass (exact match 0.949--0.961, clause-F1 0.974--0.981). On held-out controls, \Binding{} is correct in 16 of 18 seed-control cases and \Geometry{} in 22 of 27, and the direct and recalibrated labels agree on all 45 paired entries. In all six seed-by-rule runs, \Binding{} is \StateCausalUse{} and no \Geometry{} object passes closure, leaving the family \StateUnsupported{}, with no change to any family interface. Because the T5-base audit contains no class-specific \Geometry{} known positive, it establishes cross-scale verdict replication but not positive sensitivity of the \Geometry{} detectors at that scale.

For SLOG, a 96-pair exact-contrast subdomain was selected on structural grounds before any outcome was viewed, and the complete \Binding{} data--evidence--decision chain was run with three T5-small seeds. Unchanged adjudication returns \StateCausalUse{}, \StateUndetermined{}, and \StateCausalUse{}. The port required a 705-line adapter and no change to the family operations.

\FloatBarrier
\section{Supplementary Artifact Package}
\label{app:artifact_details}

The released repository contains the matched-set generator, schemas, canonical items, and validation reports; the evidence blocks, decision layer and its frozen scoring configurations, closure driver, and detector-truthfulness harness; per-run measurement summaries from the calibration, target, closure, and detector-truthfulness audits; training and audit code and runners for the robustness, truth-control, transfer, and component-rule experiments; and the inputs and script of the walkthrough of Appendix~\ref{app:walkthrough}. Its structure mirrors the paper: the matched-set files correspond to Section~\ref{sec:controlled_testbed}, the protocol code to Sections~\ref{sec:evidence} and~\ref{sec:scoring}, and the reports to Section~\ref{sec:results}. Trained checkpoints are not distributed. The testbed is produced deterministically by the released generator (seed 13), and regeneration reproduces the released files byte for byte; the generated files and a data card describing their construction, intended use, and limitations are included in the repository, and the code and data are released under the MIT license. The audits use public T5 checkpoints \citep{raffel2020t5} (Apache 2.0) and the SLOG benchmark \citep{li2023slog} (MIT license).

The examples are synthetic and contain no human-subject or personal data. The artifact is intended for auditing representation-level claims under exact matched contrasts, not for ranking general model performance.

\FloatBarrier
\section{Terminology and Artifact Mapping}
\label{app:terminology}

The paper uses descriptive names, and the released reports use shorter internal labels for the same objects. Table~\ref{tab:app_terminology} maps one to the other. The word \emph{family} is reserved for explanatory families.

\begin{table}[ht]
\centering
\small
\begin{tabular}{@{}p{0.40\linewidth}p{0.52\linewidth}@{}}
\toprule
Paper term & Artifact label \\
\midrule
\StateUnsupported{}, \StatePresentOnly{}, \StateCausalUse{}, \StateUndetermined{} & \texttt{H0}, \texttt{H1}, \texttt{H2}, \texttt{underdetermined} \\
Presence readout, interchange, necessity, repair, item linkage & Modules \texttt{M1}, \texttt{M2}, \texttt{M3}, \texttt{M4}, \texttt{M5} \\
\Binding{} and \Geometry{} families & \texttt{G1\_binding} and \texttt{G2\_distributed\_geometry} \\
Matched set & \texttt{counterpart\_family} \\
Matched easy source and hard target & \texttt{rescue\_donor} and \texttt{rescue\_target} \\
Repair & \texttt{rescue} \\
Agent replacement, theme replacement, filler replacement & \texttt{agent\_swap}, \texttt{theme\_swap}, \texttt{filler\_swap} \\
\Geometry{} Variants A--D & \Geometry{} \texttt{H1} families A--D \\
\bottomrule
\end{tabular}
\caption{Paper terms and the corresponding artifact labels.}
\label{tab:app_terminology}
\end{table}

\makeatletter\setlength{\@fptop}{0pt}\makeatother
\end{document}